\documentclass[conference]{IEEEtran}
\IEEEoverridecommandlockouts

\usepackage{cite}
\usepackage{amsmath,amssymb,amsfonts}
\usepackage{algorithmic}
\usepackage{graphicx}
\usepackage{amsmath}
\usepackage{amsthm}
\usepackage{multirow}
\usepackage{booktabs}
\usepackage{algorithm}
\usepackage{algorithmic}
\usepackage[switch]{lineno}
\usepackage{url}
\usepackage{subfigure}
\usepackage{mathtools}
\usepackage{textcomp}
\usepackage{xcolor}
\def\BibTeX{{\rm B\kern-.05em{\sc i\kern-.025em b}\kern-.08em
    T\kern-.1667em\lower.7ex\hbox{E}\kern-.125emX}}
\begin{document}


\title{MetaErr: Towards Predicting Error Patterns in Deep Neural Networks}

\author{\IEEEauthorblockN{Varun Totakura}
\IEEEauthorblockA{\textit{Department of Computer Science} \\
\textit{Florida State University}\\
Tallahassee, FL, USA \\
vt22e@fsu.edu}
\and
\IEEEauthorblockN{Shayok Chakraborty}
\IEEEauthorblockA{\textit{Department of Computer Science} \\
\textit{Florida State University}\\
Tallahassee, FL, USA \\
schakraborty2@fsu.edu}
}

\maketitle

\begin{abstract}
Due to the unprecedented success of deep learning, it has become an integral component in several multimedia computing applications in today's world. Unfortunately, deep learning systems are not perfect and can fail, sometimes abruptly, without prior warning or explanation. While reducing the error rate of deep neural networks has been the primary focus of the multimedia community, the problem of predicting when a deep learning system is going to fail has received significantly less research attention. In this paper, we propose a simple, yet effective framework, MetaErr, to address this under-explored problem in deep learning research. We train a meta-model whose goal is to predict whether a base deep neural network will succeed or fail in predicting a particular data sample, by observing the base model's performance on a given learning task. The meta-model is completely agnostic of the architecture and training parameters of the base model. Such an error prediction system can be immensely useful in a variety of smart multimedia applications. Our empirical studies corroborate the promise and potential of our framework against competing baselines. We further demonstrate the usefulness of our framework to improve the performance of pseudo-labeling based semi-supervised learning, and show that MetaErr outperforms several strong baselines on three benchmark computer vision datasets. 
\end{abstract}


\section{Introduction}

Deep learning has revolutionized the field of multimedia computing and has depicted state-of-the-art performance in a variety of applications \cite{Yoo_2019, DL_Object_Det, DeepLabV3}. Due to their remarkable performance, deep neural networks (DNNs) are now an integral component of several smart multimedia applications that involve complex decision-making, such as self-driving cars \cite{E2E_SelfDrivingCars}, smart stadiums \cite{DL_SmartStadium} and healthcare \cite{DL_Healthcare} among others. The accuracy of these systems has increased significantly over the years; however, due to the ambiguous nature of many problems, they are prone to errors and are not perfect. While the multimedia community has primarily focused on minimizing the error rates of these systems, embracing and effectively dealing with the failures has received considerably less research attention. This is very crucial, especially in safety-critical applications; if the output of a system is likely to be incorrect, it should be flagged for review by a human expert. For instance, in medical diagnosis, a human doctor should be engaged when the output of an AI system to predict the diagnosis with respect to a particular symptom is likely to be erroneous \cite{CalibratingEstimates}. Similarly, when a self-driving car's object / pedestrian detector is likely to fail, the car should rely on fallback sensors, or choose the safer action of slowing down \cite{BayesianDL}. An incorrect prediction in these applications may result in the loss of human life \cite{NHT_Safety}. 

\begin{figure*}[ht]
  \centering
   \includegraphics[width=0.75\textwidth]{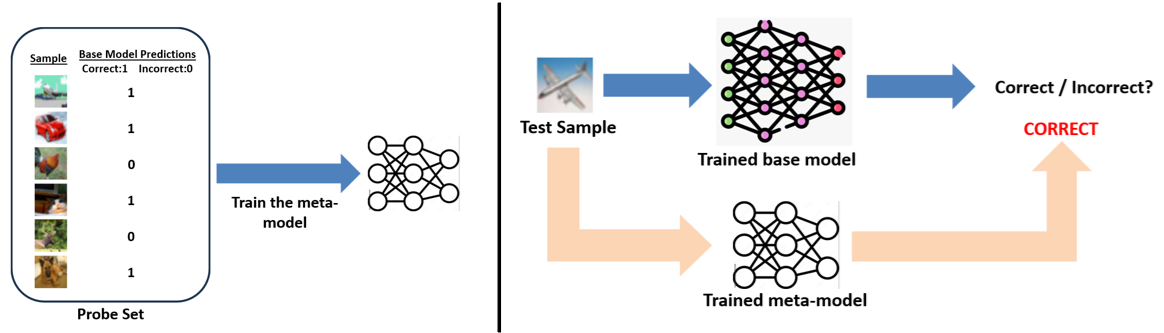}
   \caption{Outline of the proposed \textit{MetaErr} framework.}
   \label{fig_framework}
\end{figure*}

It is thus desirable to equip the base DNN model in these applications, with a capability to generate a warning when it is unable to make a reliable decision on a given test case. In this paper, we propose a simple, yet effective framework, \textit{MetaErr}, to address this practical challenge. We train a meta DNN model to predict the error patterns of the base model on a given learning task. The meta-model merely needs to have access to the predictions of the base model on a probe set (a held-out subset of the data used to train / fine-tune the base model); it needs no information about the architecture, training parameters and internal working mechanism of the base model. For a given test sample, the meta-model predicts whether the base model will classify the sample correctly or not. By identifying the examples where the base model is likely to fail, the robustness of the overall pipeline can be greatly improved. An outline of the proposed framework is depicted in Figure \ref{fig_framework}.

Further, many real-world systems are sequential in nature, where the output of one module is fed as an input to another. 
A framework like \textit{MetaErr} can be immensely useful in such systems, to improve the performance of the downstream application. 
For example, suppose a robot is classifying frames from its video feed while navigating through an unknown environment \cite{41_Parikh}. If \textit{MetaErr} can identify the frames where the classification result is likely to be incorrect, the robot can discard those frames and focus only on the frames where the classification result is predicted to be correct. This will not only enable the robot to understand the environment better, but will also be computationally more efficient (since it does not need to classify every single frame). Similarly, the output of an image aesthetic analysis module may be fed as an input to a multimedia content development system \cite{Aesthetics}. 

We validate the performance of \textit{MetaErr} on three datasets and demonstrate that our simple, yet effective system can predict errors of the base model with significant reliability. We further study the usefulness of \textit{MetaErr} to improve the performance of pseudo-labeling based semi-supervised learning, and show that it consistently outperforms several strong baselines on three benchmark datasets.


\section{Related Work}

\hspace{.2in} \textbf{Estimating Prediction Confidence of DNNs:} The confidence of a deep neural network in its decision is often correlated to the likelihood of it being correct. Reliably estimating the confidence / uncertainty of DNNs is a well-researched problem in the multimedia community \cite{BayesianDL, DeepEnsembles}. In the Monte-Carlo dropout (MC-dropout) method \cite{BayesianApproximation, Single_shot_Dropout}, dropout is used as a regularization term to compute the prediction uncertainty, and has been used in several applications. Shannon's entropy is another widely used metric to quantify the uncertainty of DNNs \cite{Entropy_1, Entropy_2}. 
The Softmax Response (SR) score proposed by Geifman and El-Yaniv \cite{SelectiveClassification} estimates the prediction confidence as the maximum softmax activation output (or the maximum class probability) furnished by a DNN for a given sample, and is also a widely used metric. 
Other methods such as Monte Carlo Batch Normalization (MCBN) \cite{BayesianUncertainty} and Bayesian methods \cite{MatrixGaussianPosteriors, BALD_Paper} have also been proposed for uncertainty estimation in DNNs. Mukhoti \textit{et al.} \cite{Mukhoti_2023_CVPR} very recently proposed the deep deterministic uncertainty (DDU) method, which uses both a discriminative classifier to capture aleatoric uncertainty and a separate feature-density estimator to capture epistemic uncertainty. Liu \textit{et al.} \cite{liu2020energy} also recently proposed an energy based score function to compute prediction uncertainty and detect out-of-distribution samples.



Many confidence estimation methods assume a ``\textit{white-box}'' setup, which has access to the architecture and the working mechanism of the base model \cite{ConfidenceScoring}. A few methods require a modification of the training algorithm of the base model, in order to jointly optimize other parameters \cite{SelectiveNet, Learning_Rejection}. In contrast, \textit{MetaErr} operates on a complete ``\textit{black-box}'' setup, where it assumes no knowledge of the architecture or the working mechanism of the base model; it also does not require any modifications to the architecture / training method of the base model, and can thus be easily integrated in several applications. 
A few methods, based on training performance predictors, require the prediction information of the base model on every test instance (such as highest and second highest predicted class confidence, relative frequency of predicted classes, entropy of the confidence vector etc.) which are used as features to predict its performance \cite{Elder_2021, Schelter_2020}. For these methods, the base model needs to be run for every test instance for its performance to be predicted for that instance. Such methods may not be efficient in situations where running the base model itself may be computationally expensive. 
In contrast, \textit{MetaErr} (once trained), only requires the test instance to predict the performance of the base model on that instance. 

\textbf{Predicting Generalization Capability of DNNs:} A body of research has focused on estimating the generalization error of DNNs on unseen test images \cite{11_in_Ranking, 38_in_Ranking}. Deng \textit{et al.} \cite{8_in_Ranking, 9_in_Ranking} proposed the AutoEval framework to automatically estimate the accuracy of DNNs on unlabeled test sets, based on their performance on a meta dataset (generated from the original dataset via various transformations). However, these methods attempt to estimate the performance of a trained DNN on a test set as a whole, rather than a single test sample, which is more challenging and is the focus of this research. 

\textbf{Failure Explainability:} In recent years, there has been a growing interest in the explainability of failures made by deep learning models \cite{Errudite, AccountableAI}. For instance, the \textit{Barlow} method, proposed by Singla \textit{et al.} \cite{UnderstandingFailures}, characterizes and explains system failures by identifying the visual attributes whose presence or absence results in poor performance. \textit{MetaErr} focuses on accurately predicting the error of a trained base DNN on a given sample, rather than the interpretability aspect of what caused the error.  


\section{Proposed Method}

We are given a base DNN model $\mathcal{F}$ which has already been trained for a given learning task $T$. Our objective is to develop a framework to identify the error / failure patterns of the base model; specifically, we are interested to quantify the performance of $\mathcal{F}$ on a given test sample $x$. We propose to incorporate a meta DNN model $\mathcal{M}$, that is trained to produce a measure of error / accuracy for each prediction made by $\mathcal{F}$. We assume a black-box setup where the meta-model only needs access to the predictions of the trained base model on a probe set; the meta-model is completely agnostic of the architecture, training parameters and the internal working mechanism of the base model. 

Formally, let $(x_{i},y_{i}),  i = 1, 2 \ldots N$ denote a probe set (a held-out validation set of the data used to train / fine-tune the base model $\mathcal{F}$), where $x_{i}$ denotes the $i^{th}$ image and $y_{i}$ denotes its corresponding ground-truth label. 
Let $(\widehat{y_{1}}, \widehat{y_{2}} \ldots \widehat{y_{N}})$ denote the predictions of the trained base model $\mathcal{F}$ on this probe set. We now describe the training mechanism of the meta-model $\mathcal{M}$. 
\noindent We define a meta-label vector $\widetilde{Y_{i}}$ as follows:
\begin{equation}
\widetilde{Y_{i}} = 
\begin{cases}
    0                          & \text{if} \hspace{.1in} y_{i} \neq \widehat{y_{i}} \\
    1                          & \text{if} \hspace{.1in} y_{i} = \widehat{y_{i}}
\end{cases}
\end{equation}
\noindent That is, $\widetilde{Y_{i}}$ is $0$ if the prediction of the base model $\mathcal{F}$ on probe image $x_{i}$ is incorrect and $1$ if it is correct. The meta-model $\mathcal{M}$ is then trained on the set $(x_{i}, \widetilde{Y_{i}}), i = 1, 2, \ldots N$. Since the meta-labels are discrete and binary, we use the binary cross entropy loss to train the meta-model:
\begin{equation}
\mathcal{L}_{meta}^{cls} = - \frac{1}{N} \sum_{i=1}^{N} \widetilde{Y_{i}}. \log (p(\widetilde{Y_{i}})) + (1 - \widetilde{Y_{i}}). \log (1 - p(\widetilde{Y_{i}}))
\end{equation}
\noindent where $p(\widetilde{Y_{i}})$ is the probability furnished by the meta-model for probe image $x_{i}$. 

We name our framework \textit{MetaErr}, as it uses a meta-model to predict the error patterns of the base model. Our empirical results corroborate that such a simple framework can be extremely useful in appropriately identifying the instances where the base model is going to fail. We note that our famework can easily be adapted to other learning tasks, such as regression and semantic segmentation by defining the meta-labels $\widetilde{Y_{i}}$ and the loss function to train the meta-model accordingly. For instance, for a regression task (where the labels $y_{i}$ are continuous), the meta-label vector can be defined as $\widetilde{Y_{i}} = \left|y_{i} - \widehat{y_{i}} \right|$, so as to capture the absolute error between the ground truth label and the prediction furnished by $\mathcal{F}$ on probe image $x_{i}$. The meta-model $\mathcal{M}$ can then be trained on the set $(x_{i}, \widetilde{Y_{i}}), i = 1, 2, \ldots N$ using the mean squared error (MSE) loss, as the meta-labels (response variable) are continuous values. For a semantic segmentation task, the meta-label vector can be defined as $\widetilde{Y_{i}} = IoU(y_{i}, \widehat{y_{i}})$, where the $IoU(:,:)$ function computes the intersection-over-union between the ground truth segmentation and the segmentation furnished by $\mathcal{F}$ on probe image $x_{i}$. The meta-model can then be trained on the set $(x_{i}, \widetilde{Y_{i}}), i = 1, 2, \ldots N$ using the MSE loss as before (as IoU is a continuous value between $0$ and $1$). 

In this paper, we only consider scenarios where $\widetilde{Y_{i}}$ is a scalar, but our method can be extended to scenarios where $\widetilde{Y_{i}}$ is a vector. Domain knowledge about the working mechanism of the base model and its specific failure modes can also be used to design features and train the meta-model accordingly. In this paper, we assume the more general case where $\mathcal{M}$ is completely agnostic of the internal working mechanism of $\mathcal{F}$. 

\textit{In our experimental studies, we used a held-out subset of the training / fine-tuning data for the base model $\mathcal{F}$ as the probe set to train the meta-model, so as to not use any additional training data, and for fair comparison with the competing baseline methods. One potential limitation of \textit{MetaErr} is that we need to train an extra meta-model, which increases the computational overhead. However, given the consistent improvements in performance achieved by our framework (as demonstrated in our experiments), we believe that this method still holds promise and potential for real-world multimedia computing applications.}


\begin{table*}[h]
\centering
\scriptsize
\caption{Mean ($\pm$ std) accuracy values (in percentage) of all the methods at each declaration rate for the \textbf{CIFAR-10} dataset. Best accuracy values at each declaration rate are marked in \textbf{bold}. Second best accuracy values are \underline{underlined}. 
}
\begin{tabular}{|c|c|c|c|c|c|c|c|c|}
\hline
\multirow{2}{*}\textbf{DR} & \textbf{Random} & \textbf{BALD \cite{BALD_Paper}} & \textbf{MC Dropout \cite{BayesianApproximation}} & \textbf{SR \cite{SelectiveClassification}} & \textbf{Energy  \cite{liu2020energy}} & \textbf{DDU \cite{Mukhoti_2023_CVPR}} & \textbf{MetaErr} & \textbf{MetaErr + SR} \\
\hline
\multicolumn{1}{|c|}{10\%} & $64.80 \pm 1.02$ & $96.77 \pm 0.61$ & $62.73 \pm 0.77$ & $\textbf{100.00} \pm 0.00$ & $85.20 \pm 0.69$ & $85.87 \pm 0.12$ & $\underline{99.93} \pm 0.09$ & $\textbf{100.00} \pm 0.00$ \\
\hline
\multicolumn{1}{|c|}{20\%} & $64.45 \pm 0.45$ & $94.05 \pm 0.41$ & $63.68 \pm 1.03$ & $\textbf{100.00} \pm 0.00$ & $76.63 \pm 0.91$ & $82.35 \pm 0.16$ & $\underline{99.97} \pm 0.05$ & $\textbf{100.00} \pm 0.00$ \\
\hline
\multicolumn{1}{|c|}{30\%} & $64.34 \pm 0.41$ & $90.70 \pm 0.52$ & $64.02 \pm 1.23$ & $\textbf{100.00} \pm 0.00$ & $75.81 \pm 0.43$ & $76.56 \pm 0.03$ & $\underline{99.98} \pm 0.03$ & $\textbf{100.00} \pm 0.00$ \\
\hline
\multicolumn{1}{|c|}{40\%} & $64.37 \pm 0.35$ & $87.13 \pm 0.28$ & $64.07 \pm 0.49$ & $\textbf{100.00} \pm 0.00$ & $65.79 \pm 0.57$ & $68.47 \pm 0.12$ & $\underline{99.98} \pm 0.02$ & $\textbf{100.00} \pm 0.00$ \\
\hline
\multicolumn{1}{|c|}{50\%} & $64.33 \pm 0.27$ & $82.88 \pm 0.39$ & $64.33 \pm 0.44$ & $99.80 \pm 0.00$ & $65.79 \pm 0.27$ & $66.18 \pm 0.10$ & $\underline{99.99} \pm 0.02$ & $\textbf{100.00} \pm 0.00$ \\
\hline
\multicolumn{1}{|c|}{60\%} & $64.17 \pm 0.18$ & $78.91 \pm 0.44$ & $64.22 \pm 0.37$ & $99.30 \pm 0.08$ & $65.51 \pm 0.16$ & $66.06 \pm 0.05$ & $\underline{99.99} \pm 0.02$ & $\textbf{100.00} \pm 0.00$ \\
\hline
\multicolumn{1}{|c|}{70\%} & $64.28 \pm 0.25$ & $75.34 \pm 0.16$ & $64.26 \pm 0.15$ & $90.87 \pm 0.05$ & $65.26 \pm 0.16$ & $65.32 \pm 0.13$ & $\underline{91.51} \pm 0.49$ & $\textbf{92.54} \pm 0.44$ \\
\hline
\multicolumn{1}{|c|}{80\%} & $64.06 \pm 0.10$ & $71.52 \pm 0.23$ & $64.23 \pm 0.32$ & $78.76 \pm 0.40$ & $64.91 \pm 0.10$ & $64.47 \pm 0.06$ & $\underline{80.07} \pm 0.42$ & $\textbf{81.10} \pm 0.38$ \\
\hline
\multicolumn{1}{|c|}{90\%} & $64.24 \pm 0.22$ & $67.96 \pm 0.20$ & $64.13 \pm 0.35$ & $70.19 \pm 0.34$ & $64.55 \pm 0.15$ & $64.38 \pm 0.02$ & $\underline{71.17} \pm 0.37$ & $\textbf{72.19} \pm 0.34$ \\
\hline
\multicolumn{1}{|c|}{100\%} & $64.08 \pm 0.30$ & $64.08 \pm 0.30$ & $64.08 \pm 0.30$ & $64.08 \pm 0.30$ & $64.08 \pm 0.30$ & $64.08 \pm 0.30$ & $64.08 \pm 0.30$ & $64.08 \pm 0.30$ \\
\hline
\end{tabular}
\label{tab_cifar10}
\end{table*}

\begin{table*}[h]
\centering
\scriptsize
\caption{Mean ($\pm$ std) accuracy values (in percentage) of all the methods at each declaration rate for the \textbf{CIFAR-100} dataset. Best accuracy values at each declaration rate are marked in \textbf{bold}. Second best accuracy values are \underline{underlined}.} 
\begin{tabular}{|c|c|c|c|c|c|c|c|c|}
\hline
\multirow{2}{*}\textbf{DR} &  \textbf{Random} &  \textbf{BALD \cite{BALD_Paper}} &  \textbf{MC Dropout \cite{BayesianApproximation}} &  \textbf{SR \cite{SelectiveClassification}} & \textbf{Energy \cite{liu2020energy}} & \textbf{DDU \cite{Mukhoti_2023_CVPR}} & \textbf{MetaErr} &  \textbf{MetaErr + SR} \\
\hline
\multicolumn{1}{|c|}{10\%} & $22.61 \pm 0.28$ & $67.93 \pm 1.30$ & $24.00 \pm 0.67$ & $\textbf{100.00} \pm 0.00$ & $68.07 \pm 2.91$ & $\underline{76.45} \pm 0.23$ & $\textbf{100.00} \pm 0.00$ & $\textbf{100.00} \pm0.00$ \\
\hline
\multicolumn{1}{|c|}{20\%} & $22.80 \pm 0.25$ & $55.50 \pm 0.40$ & $23.72 \pm 0.37$ & $\textbf{100.00} \pm 0.00$ & $67.62 \pm 1.63$ & $\underline{73.06} \pm 0.10$ & $\textbf{100.00} \pm0.00$ & $\textbf{100.00} \pm 0.00$ \\
\hline
\multicolumn{1}{|c|}{30\%} & $23.03 \pm 0.14$ & $47.82 \pm 0.35$ & $23.68 \pm 0.39$ & $76.19 \pm 0.87$ & $67.14 \pm 1.39$ & $67.23 \pm 0.62$ & $\underline{77.19} \pm 0.87$ & $\textbf{78.15} \pm 0.61$ \\
\hline
\multicolumn{1}{|c|}{40\%} & $23.43 \pm 0.39$ & $41.66 \pm 0.37$ & $23.10 \pm 0.57$ & $56.89 \pm 0.66$ & $58.58 \pm 1.40$ & $\textbf{59.89} \pm 0.52$ & $57.89 \pm 0.66$ & $\underline{58.62} \pm 0.91$ \\
\hline
\multicolumn{1}{|c|}{50\%} & $23.51 \pm 0.34$ & $36.77 \pm 0.08$ & $23.03 \pm 0.38$ & $45.11 \pm 0.28$ & $\textbf{48.36} \pm 1.33$ & $43.10 \pm 0.90$ & $46.31 \pm 0.52$ & $\underline{46.86} \pm 0.55$ \\
\hline
\multicolumn{1}{|c|}{60\%} & $23.40 \pm 0.13$ & $32.95 \pm 0.14$ & $23.17 \pm 0.20$ & $37.43 \pm 0.44$ & $\textbf{39.22} \pm 1.22$ & $37.43 \pm 0.82$ & $38.59 \pm 0.43$ & $\underline{39.19} \pm 0.54$ \\
\hline
\multicolumn{1}{|c|}{70\%} & $23.12 \pm 0.15$ & $29.71 \pm 0.22$ & $23.15 \pm 0.12$ & $32.18 \pm 0.43$ & $33.07 \pm 0.77$ & $\underline{33.11} \pm 0.19$ & $33.08 \pm 0.37$ & $\textbf{33.71} \pm 0.39$ \\
\hline
\multicolumn{1}{|c|}{80\%} & $23.17 \pm 0.22$ & $27.18 \pm 0.19$ & $23.32 \pm 0.16$ & $28.11 \pm 0.45$ & $27.97 \pm 0.51$ & $\textbf{30.67} \pm 0.24$ & $28.94 \pm 0.33$ & $\underline{29.77} \pm 0.48$ \\
\hline
\multicolumn{1}{|c|}{90\%} & $23.16 \pm 0.23$ & $24.96 \pm 0.23$ & $23.24 \pm 0.30$ & $24.96 \pm 0.53$ & $24.84 \pm 0.46$ & $23.58 \pm 0.07$ & $\underline{25.73} \pm 0.29$ & $\textbf{26.46} \pm 0.45$ \\
\hline
\multicolumn{1}{|c|}{100\%} & $23.16 \pm 0.26$ & $23.16 \pm 0.26$ & $23.16 \pm 0.26$ & $23.16 \pm 0.26$ & $23.16 \pm 0.26$ & $23.16 \pm 0.26$ & $23.16 \pm 0.26$ & $23.16 \pm 0.26$ \\
\hline
\end{tabular}
\label{tab_cifar100}
\end{table*}

\begin{table*}[h]
\centering
\scriptsize
\caption{Mean ($\pm$ std) accuracy values (in percentage) of all the methods at each declaration rate for the \textbf{SVHN} dataset. Best accuracy values at each declaration rate are marked in \textbf{bold}. Second best accuracy values are \underline{underlined}. }
\begin{tabular}{|c|c|c|c|c|c|c|c|c|}
\hline
\multirow{2}{*}\textbf{DR} &  \textbf{Random} &  \textbf{BALD \cite{BALD_Paper}} &  \textbf{MC Dropout \cite{BayesianApproximation}} &  \textbf{SR \cite{SelectiveClassification}} & \textbf{Energy \cite{liu2020energy}} & \textbf{DDU \cite{Mukhoti_2023_CVPR}} & \textbf{MetaErr} &  \textbf{MetaErr + SR} \\
\hline
\multicolumn{1}{|c|}{10\%} & $19.82 \pm 0.29$ & $\underline{89.30} \pm 2.32$ & $18.79 \pm 0.77$ & $\textbf{100.00} \pm 0.00$ & $79.62 \pm 1.55$ & $82.01 \pm 1.06$ & $\textbf{100.00} \pm0.00$ & $\textbf{100.00} \pm0.00$ \\
\hline
\multicolumn{1}{|c|}{20\%} & $20.14 \pm 0.19$ & $79.89 \pm 1.48$ & $18.83 \pm 0.77$ & $96.46 \pm 1.80$ & $78.80 \pm 1.06$ & $77.18 \pm 2.22$ & $\underline{97.79} \pm 1.79$ & $\textbf{98.39} \pm 1.67$ \\
\hline
\multicolumn{1}{|c|}{30\%} & $19.91 \pm 0.39$ & $59.97 \pm 1.18$ & $19.01 \pm 0.62$ & $63.19 \pm 2.01$ & $\textbf{68.99} \pm 1.17$ & $\underline{68.63} \pm 0.98$ & $65.19 \pm 1.20$ & $66.53 \pm 1.34$ \\
\hline
\multicolumn{1}{|c|}{40\%} & $19.75 \pm 0.45$ & $40.22 \pm 1.18$ & $19.04 \pm 0.56$ & $47.23 \pm 1.06$ & $\underline{48.98} \pm 0.91$ & $46.98 \pm 0.18$ & $48.89 \pm 0.90$ & $\textbf{51.23} \pm 1.30$ \\
\hline
\multicolumn{1}{|c|}{50\%} & $19.61 \pm 0.46$ & $30.04 \pm 0.94$ & $19.06 \pm 0.46$ & $37.45 \pm 0.34$ & $38.93 \pm 0.71$ & $\underline{39.87} \pm 1.20$ & $39.11 \pm 0.72$ & $\textbf{40.78} \pm 1.57$ \\
\hline
\multicolumn{1}{|c|}{60\%} & $19.61 \pm 0.34$ & $24.78 \pm 1.32$ & $19.17 \pm 0.44$ & $31.73 \pm 0.65$ & $30.99 \pm 0.63$ & $30.12 \pm 0.69$ & $\underline{32.59} \pm 0.60$ & $\textbf{34.59} \pm 0.60$ \\
\hline
\multicolumn{1}{|c|}{70\%} & $19.57 \pm 0.33$ & $23.28 \pm 1.59$ & $19.11 \pm 0.41$ & $25.60 \pm 0.31$ & $24.10 \pm 0.54$ & $23.18 \pm 0.54$ & $\underline{27.94} \pm 0.51$ & $\textbf{28.94} \pm 0.51$ \\
\hline
\multicolumn{1}{|c|}{80\%} & $19.54 \pm 0.27$ & $21.10 \pm 0.62$ & $19.15 \pm 0.30$ & $23.11 \pm 0.56$ & $22.13 \pm 0.42$ & $22.88 \pm 0.39$ & $\underline{24.44} \pm 0.45$ & $\textbf{24.81} \pm 0.46$ \\
\hline
\multicolumn{1}{|c|}{90\%} & $19.54 \pm 0.35$ & $20.03 \pm 0.55$ & $19.16 \pm 0.24$ & $21.26 \pm 0.24$ & $21.14 \pm 0.33$ & $20.45 \pm 0.13$ & $\underline{21.73} \pm 0.40$ & $\textbf{22.17} \pm 0.25$ \\
\hline
\multicolumn{1}{|c|}{100\%} & $19.55 \pm 0.36$ & $19.55 \pm 0.36$ & $19.55 \pm 0.36$ & $19.55 \pm 0.36$ & $19.55 \pm 0.36$ & $19.55 \pm 0.36$ & $19.55 \pm 0.36$ & $19.55 \pm 0.36$ \\
\hline
\end{tabular}
\label{tab_svhn}
\end{table*}

\section{Experiments and Results}
\label{sec_expt}

\textbf{Datasets:} We used three benchmark computer vision datasets for our experiments: $(i)$ \textit{CIFAR-10} \cite{CIFAR_dataset}; $(ii)$ \textit{CIFAR-100} \cite{CIFAR_dataset}; and $(iii)$ \textit{SVHN} \cite{SVHN_dataset}. 

\textbf{Evaluation Metric:} We used the \textit{Accuracy vs. Declaration Rate (ADR)} metric to study the performance of our framework \cite{Failures_VisionSystems}. Declaration Rate (DR) is the percentage of test samples on which the base model outputs a prediction; for the other samples, no prediction is provided. The samples in the test set were sorted in order of decreasing confidence (\textit{MetaErr} and the baselines detailed below will each sort the test set in a different order). At a given declaration rate $d$, the top $d\%$ of the test samples in the sorted order were retreived and the accuracy of the base model only on this set was computed. A method furnishing a higher value of this accuracy is a better method, as it can more accurately identify the samples that will be predicted correctly by the base model. We used $10$ values of the DR from $10\%$ to $100\%$, in steps of $10\%$. 

\textbf{Experimental Setup:} After training the base model $\mathcal{F}$, it was applied on the test set to generate a prediction for each sample. The meta-model $\mathcal{M}$, trained on the probe set, was also applied on the test set to predict which of the test samples will be correctly / incorrectly predicted by $\mathcal{F}$. The test samples were sorted in decreasing order of the probability of correct prediction,  furnished by the meta-model (that is, test samples where the meta-model $\mathcal{M}$ had higher confidence that they will be correctly predicted by the base model were ranked higher). This sorted list was then used to derive the accuracy values for ADR computation. 
Our base model $\mathcal{F}$ was trained to have low accuracy, in order to assess the error prediction performance of \textit{MetaErr} in the challenging setup where the base model is error-prone. 

\textbf{Implementation Details:} \textbf{Base Model $\mathcal{F}$:} We used the GoogleNet architecture as the base model due to its impressive performance in image classification tasks. Each convolutional layer included rectified linear unit (\textit{ReLU}) activation, together with batch normalization (\textit{BatchNorm}) for enhanced training stability. A dropout layer was included, with rate $0.2$. We used the Stochastic Gradient Descent (SGD) optimizer, and the model was trained for $50$ epochs with a learning rate of $10^{-3}$, weight decay of $5 \times 10^{-4}$ and a momentum of $0.9$. \textbf{Meta-Model $\mathcal{M}$:} The ResNet50 architecture was used as the meta-model. For the linear softmax layer, we used $2$ units, aligning with the binary nature of the network's purpose to ascertain the base model's predictions across two classes. Each convolutional layer was provided with rectified linear unit (\textit{ReLU}) activation, together with batch normalization (\textit{BatchNorm}) for enhanced training stability. We used the SGD optimizer, and the model was trained for $5$ epochs, with a learning rate at $10^{-3}$, weight decay of $5 \times 10^{-2}$ and a momentum of $0.7$. 

\textbf{Comparison Baselines:} Any strategy to sort the test samples in a particular order can be used as a comparison baseline. We used commonly used and state-of-the-art uncertainty estimation methods as comparison baselines in our work, where test samples are sorted from lowest to highest prediction uncertainties (uncertainty quantification is the most common method to assess the relibaility of a DNN's predictions): $(i)$ \textbf{Random}, where the samples in the test set were sorted randomly; $(ii)$ \textbf{BALD} \cite{BALD_Paper}; $(iii)$ \textbf{MC Dropout} \cite{BayesianApproximation}; $(iv)$ \textbf{Softmax Response (SR)} \cite{SelectiveClassification}; $(v)$ \textbf{Energy} \cite{liu2020energy}; and $(vi)$ \textbf{DDU} \cite{Mukhoti_2023_CVPR}. All these techniques have been extensively used to estimate the confidence of a DNN's predictions; moreover, \textit{Energy} and \textit{DDU} are two very recently proposed methods for uncertainty estimation in DNNs. All the results were averaged over $3$ runs to rule out the effects of randomness. 

\textbf{Results:} The results are depicted in Tables \ref{tab_cifar10}, \ref{tab_cifar100} and \ref{tab_svhn}. At a given declaration rate of $d\%$, the top $d\%$ of the test samples sorted by a given method are retrieved and the accuracy of the base model (in $\%$) on this set is depicted in the corresponding column. \textit{Random} depicts more or less the same accuracy at all values of DR, which is much less than the other methods. This denotes that if the error prediction system performs at a chance level, the accuracy will be poor (as expected) and on an average, will remain constant with varying DR. For all the other methods, the accuracy depicts a decreasing trend with increasing DR. This is because, as the DR increases, the system is expected to make more and more predictions on the test set, which increases its chances of making incorrect predictions. The other baselines depict much better error prediction performance than \textit{Random}, as their belief is based on the probability distribution furnished by $\mathcal{F}$ on a given test sample. 
The performance of \textit{MC Dropout} is more or less similar to that of \textit{Random}. The \textit{Energy} and \textit{DDU} methods depict good performance for the CIFAR-100 dataset and obtain the highest accuracy for $4$ DR values; however, they are not consistent across datasets in their performance. \textit{BALD} outperforms \textit{Energy} and \textit{DDU} for some of the low DR values in CIFAR-10 and SVHN. \textit{MetaErr} consistently depicts impressive performance across all the datasets. For most of the DR values, it produces the highest accuracy compared to all the baselines, across all the datasets. 
This shows that \textit{MetaErr} can accurately identify the test samples where the base model's predictions are likely to be incorrect; thus, it can be used to filter out the unreliable samples, and improve the overall performance of a system. At the declaration rate of $100\%$, a prediction is provided for all the test samples and the sorting order does not matter. All the methods therefore depict the same accuracy. 

The baseline methods are mostly based on uncertainty estimation of the base model $\mathcal{F}$. 
Modern DNNs are often known to be miscalibrated and their prediction confidence do not always bear a strong correlation with their probability of failure on a given image \cite{MixupTraining, Calibration_NN}. Thus, these confidence / uncertainty measures, which are derived using the softmax probabilities may not be reliable always. \textit{MetaErr} on the other hand, is directly trained to capture the error patterns of $\mathcal{F}$ and thus depicts much better performance than these baselines. A combination of \textit{MetaErr} and \textit{SR} (by a simple averaging of the respective probabilities) depicts the best results, as evident from Tables \ref{tab_cifar10}, \ref{tab_cifar100} and \ref{tab_svhn}. Thus, \textit{MetaErr} captures orthogonal properties as the base model's confidence and combining them further improves the results. These results depict the promise and potential of \textit{MetaErr} for smart multimedia applications.

\begin{figure}[h]
  \centering
   \includegraphics[width=0.85\linewidth]{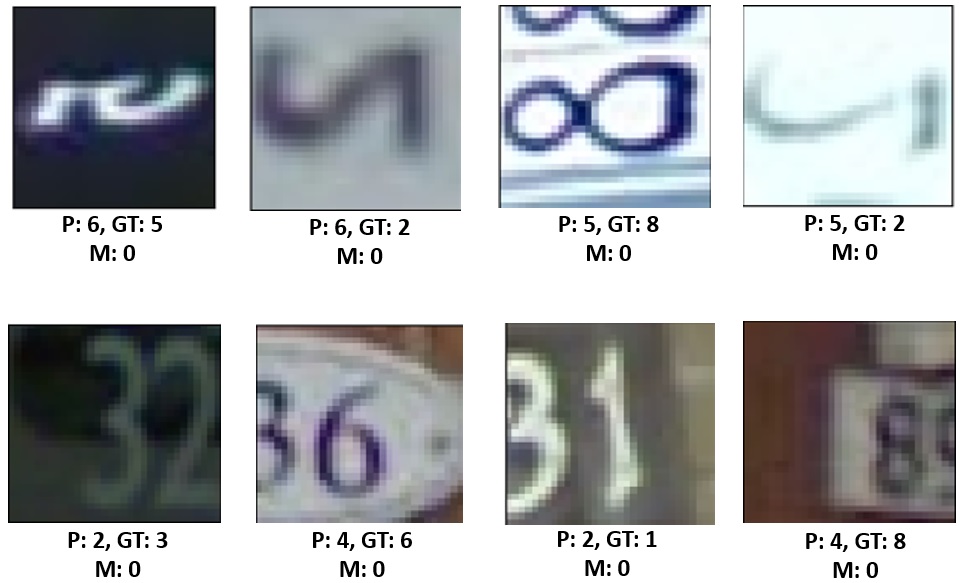}
   \caption{Visual illustration of the performance of \textit{MetaErr} on the SVHN dataset. Here, \textbf{P} denotes the class predicted by the base model $\mathcal{F}$, \textbf{GT} denotes the ground-truth class and \textbf{M: 0} denotes that the meta-model $\mathcal{M}$ predicted that the base model will commit an error.}
   \label{fig_visual}
\end{figure}

\textbf{Visual Illustrations:} A visual illustration of the performance of \textit{MetaErr} on the SVHN dataset is depicted in Figure \ref{fig_visual}. Here, \textbf{P} denotes the class predicted by the base model $\mathcal{F}$, \textbf{GT} denotes the ground-truth class and \textbf{M: 0} denotes that the meta-model $\mathcal{M}$ predicted that the base model will commit an error. We selected a few sample images from the test set, where the digit was flipped in a horizontal position (top row) or where there were more than one digit in the image (bottom row). These are challenging conditions (as evident visually), which make it difficult for the base model to make a correct prediction. In all these cases, the meta-model predicted that the base model will commit an erroneous prediction on the image, which was really the case. This once again reinforces the efficacy of the meta-model to identify the samples where the base model is likely to fail. 

\begin{table*}[h]
\centering
\scriptsize
\caption{Mean ($\pm$ std) accuracy values (in percentage) of all the SSL methods with $10\%$ labeled training data for all the datasets. Best accuracy values are marked in \textbf{bold}. }
\label{SSL_default}
\begin{tabular}{|c|c|c|c|c|c|c|}
\hline
\multirow{2}{*}{Method} & \multicolumn{2}{c|}{\textbf{FMNIST}} & \multicolumn{2}{c|}{\textbf{CIFAR 10}} & \multicolumn{2}{c|}{\textbf{SVHN}} \\
\cline{2-7}
& \textbf{Top-1} & \textbf{Top-5} & \textbf{Top-1} & \textbf{Top-5} & \textbf{Top-1} & \textbf{Top-5}\\
\hline
\multicolumn{1}{|c|}{\textbf{ICT \cite{ICT}}} & $72.57 \pm 0.95$ & $97.94 \pm 0.22$ & $38.63 \pm 1.48$ & $89.16 \pm 0.63$ & $20.41 \pm 0.93$ & $67.21 \pm 1.36$ \\
\hline
\multicolumn{1}{|c|}{\textbf{VAT \cite{VAT}}} & $72.09 \pm 1.59$ & $98.13 \pm 0.10$ & $41.44 \pm 1.74$ & $89.33 \pm 1.25$ & $21.13 \pm 1.09$ & $54.04 \pm 1.57$ \\
\hline
\multicolumn{1}{|c|}{\textbf{LP \cite{LabelPropagation}}} & $71.62 \pm 2.39$ & $97.38 \pm 0.49$ & $43.23 \pm 0.66$ & $90.12 \pm 0.71$ & $19.71 \pm 0.14$ & $64.52 \pm 2.21$ \\
\hline
\multicolumn{1}{|c|}{\textbf{PL-CB \cite{PL_CB}}} & $73.12 \pm 3.17$ & $97.71 \pm 0.17$ & $44.51 \pm 2.50$ & $90.41 \pm 0.80$ & $19.04 \pm 0.87$ & $62.88 \pm 2.29$ \\
\hline
\multicolumn{1}{|c|}{\textbf{CL \cite{CurriculumLabeling}}} & $72.69 \pm 0.73$ & $97.86 \pm 0.28$ & $45.55 \pm 0.64$ & $91.03 \pm 0.94$ & $19.47 \pm 1.32$ & $64.24 \pm 4.86$ \\
\hline
\multicolumn{1}{|c|}{\textbf{MetaErr}} & $\textbf{75.05} \pm 0.84$ & $\textbf{98.49} \pm 0.12$ & $\textbf{47.42} \pm 1.01$ & $\textbf{91.39} \pm 1.14$ & $\textbf{22.83} \pm 2.41$ & $\textbf{70.13} \pm 2.64$ \\
\hline
\end{tabular}
\end{table*}

\begin{table*}[h]
\centering
\scriptsize
\caption{Mean ($\pm$ std) accuracy values (in percentage) of all the SSL methods with $5\%$ and $1\%$ labeled training data for the \textbf{CIFAR-10} dataset. Best accuracy values are marked in \textbf{bold}.}
\begin{tabular}{|c|c|c|c|c|c|}
\hline
\multirow{2}{*}{Method} & \multicolumn{2}{c|}{\textbf{5\% Labeled Training Data}} & \multicolumn{2}{c|}{\textbf{1\% Labeled Training Data}} \\
\cline{2-5}
& \textbf{Top-1} & \textbf{Top- 5} & \textbf{Top-1} & \textbf{Top-5} \\
\hline
\multicolumn{1}{|c|}{\textbf{ICT \cite{ICT}}} & $35.63 \pm 0.89$ & $82.87 \pm 0.31$ & $15.40 \pm 0.54$ & $61.55 \pm 2.77$ \\
\hline
\multicolumn{1}{|c|}{\textbf{VAT \cite{VAT}}} & $32.30 \pm 0.74$ & $82.63 \pm 0.84$ & $13.44 \pm 1.89$ & $59.72 \pm 2.62$ \\
\hline
\multicolumn{1}{|c|}{\textbf{LP \cite{LabelPropagation}}} & $28.03 \pm 0.59$ & $77.41 \pm 1.63$ & $14.49 \pm 1.21$ & $57.57 \pm 1.57$ \\
\hline
\multicolumn{1}{|c|}{\textbf{PL-CB \cite{PL_CB}}} & $31.57 \pm 1.14$ & $82.24 \pm 0.58$ & $13.37 \pm 0.48$ & $59.18 \pm 1.27$ \\
\hline
\multicolumn{1}{|c|}{\textbf{CL \cite{CurriculumLabeling}}} & $35.66 \pm 0.61$ & $83.06 \pm 0.27$ & $\textbf{17.01} \pm 0.40$ & $70.54 \pm 0.58$ \\
\hline
\multicolumn{1}{|c|}{\textbf{MetaErr}} & $\textbf{36.25} \pm 0.93$ & $\textbf{85.28} \pm 1.22$ & $16.95 \pm 0.76$ & $\textbf{70.77} \pm 0.45$ \\
\hline
\end{tabular}
\label{SSL_less}
\end{table*}

\section{MetaErr for Semi-supervised Learning}
\label{sec_SSL}

\textit{Semi-supervised Learning (SSL)} techniques utilize large amounts of unlabeled data, in addition to a small amount of labeled data, to train a deep neural network, and have achieved remarkable performance \cite{MeanTeachers, VAT, ICT}. One of the popular SSL strategies is based on \textit{pseudo-labeling}, where unlabeled samples are iteratively added into the training set by pseudo-labeling them with a weak model trained on a combination of labeled and pseudo-labeled samples \cite{LabelPropagation, PL_CB}. Recently, self-paced curriculum labeling has been exploited for pseudo-labeling, and has depicted impressive results \cite{CurriculumLabeling}.  

Formally, in an SSL setup, we are given labeled training data $D_{L} = \{(x,y) | x \in X, y \in Y\}$ and unlabeled training data $D_{U} = \{x | x \in X\}$, where $x$ and $y$ denote the image features and the labels respectively. Typically, $|D_{L}| \ll |D_{U}|$. In the curriculum learning based pseudo-labeling framework \cite{CurriculumLabeling}, the model goes through multiple rounds of training by utilizing its own past feedback. In the first round, the model is trained on the labeled set $D_{L}$. In training round $t$, let $(X_{t}, Y_{t})$ denote the labeled training set and $\Theta_{t}$ denote the trained model parameters, where $t = {1,2, \ldots T}$. In round $t+1$, a subset of the unlabeled set $\overline{X_{t}}$ is appended to the labeled set, with labels $\overline{Y_{t}}: X_{t+1} = X_{t} \cup \overline{X_{t}}$ and $Y_{t+1} = Y_{t} \cup \overline{Y_{t}}$. Here, $\overline{Y_{t}}$ denotes the pseudo-labels of $\overline{X_{t}}$, predicted by the current model $\Theta_{t}$. In this pipeline, the output of the pseudo-labeling module is fed as an input to the model training module, to induce the deep model for the next round. 

The key to this algorithm is deciding which unlabeled samples should be appended to the training set in each round. 
\textit{MetaErr} can be leveraged in this setup to identify the unlabeled samples $\overline{X_{t}}$, whose pseudo-labels, $\overline{Y_{t}}$, are most likely predicted correctly in each training round. This will result in a better deep model for the subsequent round, which will be able to assign pseudo-labels to the remaining unlabeled samples more accurately. At the end of $T$ rounds, \textit{MetaErr} can potentially produce a deep model with better generalization capability. By appropriately identifying the samples with correctly assigned pseudo-labels, our framework can thus influence the performance of a downstream task which uses these pseudo-labels to update the deep model. 

To study this, we conducted experiments on three benchmark vision datasets: \textit{FMNIST} \cite{FMNIST_dataset}, \textit{CIFAR-10} \cite{CIFAR_dataset} and \textit{SVHN} \cite{SVHN_dataset}. For each dataset, we used $10\%$ of the provided training data as the labeled set $D_{L}$ and the remaining $90\%$ as the unlabeled set $D_{U}$ ($|D_{L}| \ll |D_{U}|$). The performance was evaluated on the test set provided with each dataset. The number of samples in the labeled, unlabeled and test sets are shown in Table \ref{tab_SSL_data}. All the results were averaged over $3$ runs to rule out the effects of randomness. 

\begin{table}[h]
	\centering
	\scriptsize
	
\caption{Number of samples in the labeled, unlabeled and test sets for the SSL experiment.}	

\begin{tabular}{|c|c|c|c|c|}
\hline
\textbf{} & \textbf{Labeled ($D_{L}$)} & \textbf{Unlabeled ($D_{U}$)} & \textbf{Test}   \\
\hline
\textbf{FMNIST} & $6,000$ & $54,000$ & $10,000$  \\
\hline
\textbf{CIFAR-10} & $5,000$ & $45,000$ & $10,000$   \\
\hline
\textbf{SVHN} & $7,325$ & $65,932$ & $26,032$  \\
\hline

\end{tabular}

	\label{tab_SSL_data}
\end{table}

\textbf{Comparison Baselines:} We compared the performance of \textit{MetaErr} against several recent deep SSL techniques: $(i)$ \textit{Curriculum Labeling (CL)} \cite{CurriculumLabeling};  
$(ii)$ \textit{Pseudo-Labeling with Confirmation Bias (PL-CB)} \cite{PL_CB}; $(iii)$ \textit{Label Propagation (LP)} \cite{LabelPropagation}; $(iv)$ \textit{Interpolation Consistency Training (ICT)} \cite{ICT}; and $(v)$ \textit{Virtual Adversarial Training (VAT)} \cite{VAT}. \textbf{\textit{Note that, for our method, since we need to train the meta-model, we subdivide the labeled set $D_{L}$ into two equal parts to train the base model and the meta-model, so that the total amount of training data used remains the same as the other methods, for fair comparison.}} \textit{CL, PL-CB} and \textit{LP} are based on the idea of pseudo-labels, while \textit{ICT} and \textit{VAT} are based on consistency regularization. Consistency regularization is another popular method for SSL, where a network is trained to make consistent predictions in response to perturbation of unlabeled samples, by combining the standard cross-entropy loss with a consistency loss \cite{MeanTeachers}.  


\textbf{Results:} The results on the FMNIST, CIFAR-10 and SVHN datasets are reported in Table \ref{SSL_default}. We note that \textit{MetaErr} consistently outperforms all the baselines across all three datasets, both in terms of the Top-1 and Top-5 accuracy. 
Thus, the meta-model in our framework can appropriately identify the samples that are likely to be predicted correctly by the base model; using these samples together with their predicted pseudo-labels can produce a better deep model in the subsequent SSL round, which improves the overall SSL performance. This corroborates the usefulness of \textit{MetaErr} for pseudo-labeling based SSL. 

We conducted two additional experiments where we reduced the size of the labeled set to $5\%$ and $1\%$ of the provided training set; the size of the unlabeled set was increased to $95\%$ and $99\%$ respectively. This is a typical setting in SSL, where we are given a small amount of labeled data, and a large amount of unlabeled data. The results on the CIFAR-10 dataset (averaged over $3$ random runs) are depicted in Table \ref{SSL_less}. \textit{MetaErr} once again depicts very impressive performance and attains the highest Top-1 and Top-5 accuracy with $5\%$ labeled training data and the highest Top-5 accuracy with $1\%$ labeled training data, compared to all the baselines. This shows the efficacy of our framework for pseudo-labeling based SSL in the challenging setting of extremely limited labeled samples.

\section{Conclusion}
\label{sec_conc}
 
In this paper, we proposed \textit{MetaErr}, a framework to predict the performance of a trained deep neural network on a given test sample. Our rationale was to train a meta deep neural network to model the error patterns of the base DNN. The meta-model merely needs to have access to the predictions of the base model on a probe set, and is completely agnostic to the architecture and training mechanism of the base model. Such a simple (yet effective) system can be immensely useful in several multimedia computing applications, as it can provide us insights on when the control should be passed to a human expert to analyze the predictions furnished by an AI system. We validated the performance of our framework on three benchmark datasets with promising empirical results. \textit{MetaErr} also depicted promise in improving the performance of pseudo-labeling based SSL, and consistently outperformed several strong baselines. As part of future work, we will study the performance of \textit{MetaErr} on other multimedia problems, such as regression, semantic segmentation and object detection. We will also study the performance of our framework to predict the error patterns of more advanced models such as vision transformers (ViTs).

\section{Acknowledgment}
\label{sec_ack}

This research was supported in part by the National Science Foundation under Grant Number: IIS-2143424 (NSF CAREER Award).

%

%

\bibliographystyle{IEEEtran}
\bibliography{ICSM_2025_MetaErr}

\section{Supplemental File\\}
\maketitle

\begin{abstract}
In this Supplemental File, we include results of our \textit{MetaErr} framework for regression and semantic image segmentation applications.

\end{abstract}

\begin{IEEEkeywords}
Deep Learning, Error Prediction, Multimedia
Computing.
\end{IEEEkeywords}

\subsection{MetaErr for Regression and Image Segmentation}
As detailed in the main paper, let $(x_{i},y_{i}),  i = 1, 2 \ldots N$ denote a probe set (a held-out validation set of the data used to train / finetune the base model $\mathcal{F}$), where $x_{i}$ denotes the $i^{th}$ image and $y_{i}$ denotes its corresponding ground-truth label. The nature of the labels $y_{i}$ depends on the learning task $T$; for instance, $y_{i}$ will be discrete for classification, continuous for regression, a segmented image for image segmentation and so on. Let $(\widehat{y_{1}}, \widehat{y_{2}} \ldots \widehat{y_{N}})$ denote the predictions of the trained base model $\mathcal{F}$ on this probe set. We now describe the training mechanism of the meta-model $\mathcal{M}$ for the regression and image segmentation setup.

\subsubsection{Regression Setup}

\noindent We define a meta-label vector $\widetilde{Y_{i}}$ as follows:
\begin{equation}
\widetilde{Y_{i}} = \left|y_{i} - \widehat{y_{i}} \right|
\end{equation}
\noindent That is, $\widetilde{Y_{i}}$ captures the absolute error between the ground truth label and the prediction furnished by $\mathcal{F}$ on probe image $x_{i}$. The meta-model $\mathcal{M}$ is then trained on the set $(x_{i}, \widetilde{Y_{i}}), i = 1, 2, \ldots N$. Since the meta-labels are continuous, we use the mean squared error (MSE) loss to train the meta-model in this case:
\begin{equation}
\mathcal{L}_{meta}^{reg} = \frac{1}{N} \sum_{i=1}^{N} (\widetilde{Y_{i}} - y_{i}^{meta})^{2}
\end{equation}
\noindent where  $y_{i}^{meta}$ is the prediction of the error furnished by the meta-model for probe image $x_{i}$.

\subsubsection{Image Segmentation Setup}

\noindent We define a meta label vector $\widetilde{Y_{i}}$ as follows:
\begin{equation}
\widetilde{Y_{i}} = IoU(y_{i}, \widehat{y_{i}})
\end{equation}

\noindent where the $IoU(:,:)$ function computes the intersection-over-union between the ground truth segmentation and the segmentation furnished by $\mathcal{F}$ on probe image $x_{i}$. The meta-model $\mathcal{M}$ is then trained on the set $(x_{i}, \widetilde{Y_{i}}), i = 1, 2, \ldots N$. Since the meta-labels are continuous (IoU is a continuous value between $0$ and $1$), we use the mean squared error (MSE) loss to train the meta-model, as before:
\begin{equation}
\mathcal{L}_{meta}^{seg} = \frac{1}{N} \sum_{i=1}^{N} (\widetilde{Y_{i}} - y_{i}^{meta})^{2}
\end{equation}
\noindent where  $y_{i}^{meta}$ is the prediction of the IoU furnished by the meta-model for probe image $x_{i}$. 

\subsection{Experiments and Evaluations}
\label{sec_expt}

As in the main paper, we used the \textit{Accuracy vs. Declaration Rate (ADR)} metric to study the performance of our framework. 

\subsubsection{Regression}

\hspace{.2in} \textbf{Datasets:} We used two age estimation datasets for the regression setup: $(i)$ \textbf{AgeDB} \cite{AgeDB_dataset}; and $(ii)$ \textbf{UTKFace} \cite{UTKFace_dataset}. 

\textbf{Comparison Baselines:} We used \textit{Random} and \textit{MC-Dropout} \cite{BayesianApproximation} as the baselines in this experiment. \textit{MC-Dropout} is widely used in regression applications. 

\textbf{Implementation Details:} \textbf{Base Model $\mathcal{F}$:} We used the GoogleNet architecture as the base model. Each convolutional layer included \textit{ReLU} activation, together with \textit{BatchNorm} for enhanced training stability. A dropout layer was included, with rate $0.2$. We used the SGD optimizer, and the model was trained for $15$ epochs with a learning rate of $10^{-5}$, weight decay of $10^{-5}$ and a momentum of $0.7$. \textbf{Meta-Model $\mathcal{M}$:} The ResNet50 architecture was used as the meta-model. The linear layer had a single neuron, as we are trying to predict a continuous value (absolute error of the base model). Each convolutional layer was provided with \textit{ReLU} activation, together with \textit{BatchNorm}. We used the SGD optimizer, and the model was trained for $5$ epochs, with a learning rate at $10^{-3}$, weight decay of $10^{-5}$ and a momentum of $0.9$.

\begin{figure}[h]
	\centering
		\subfigure[AgeDB]{
          \label{fig_agedb}
          \includegraphics[trim = 1.3in 3.2in 1.7in 3.4in,clip,width=.38\textwidth]{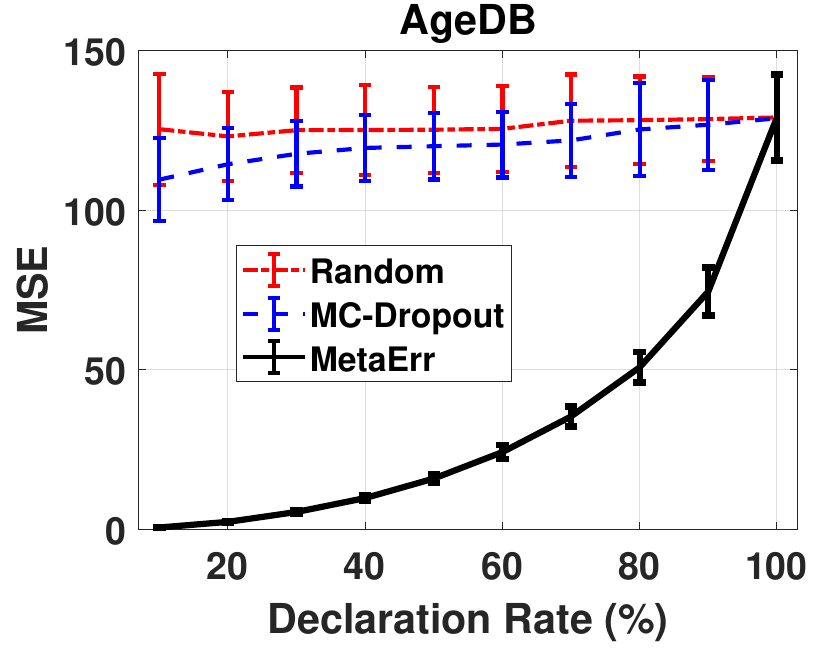}} 
     \hspace{.03in}    
     \subfigure[UTK Face]{
          \label{fig_utkface}
          \includegraphics[trim = 1.3in 3.2in 1.7in 3.4in,clip,width=.38\textwidth]{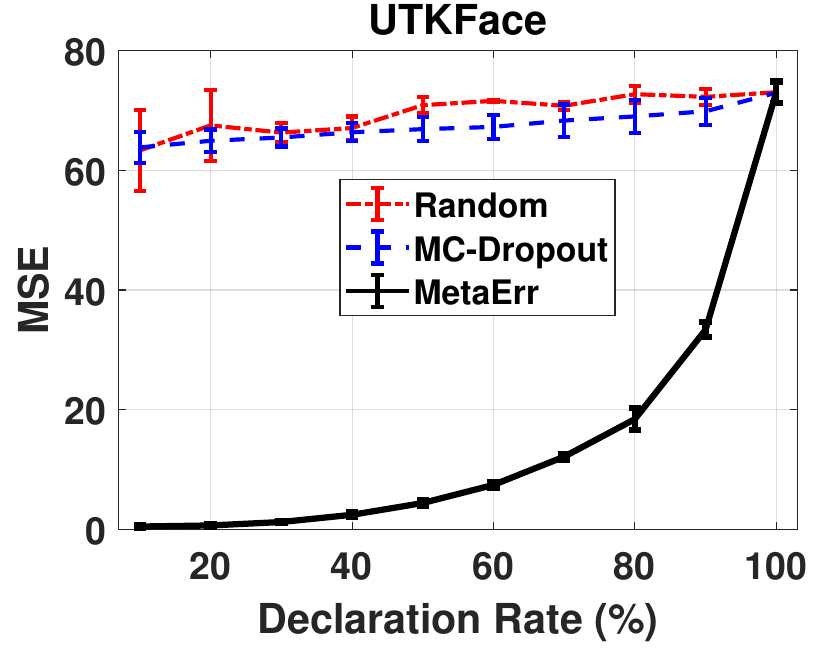}}           		
		 		
	\caption{Performance of \textit{MetaErr} on regression tasks. Best viewed in color.}
	\label{fig_reg_results}
\end{figure} 

\textbf{Results:} The results are depicted in Figure \ref{fig_reg_results}, where the $x$-axis denotes the declaration rate (DR), and the $y$-axis denotes the mean squared error (MSE). We note that for both datasets, \textit{MetaErr} depicts very low errors at low DR, corroborating its capability to appropriately identify the test samples which will be predicted with low error by the base model. \textit{Random} depicts much high error, which remains more or less constant with increasing DR. \textit{MC-Dropout} also depicts high error values (although better than \textit{Random}), once again asserting the miscalibration of prediction confidence of DNNs. \textit{MetaErr} is trained to directly model the error patterns of the base model $\mathcal{F}$ and can thus appropriately identify the images where the based model will exhibit good performance (low error). At any given DR, \textit{MetaErr} furnishes a much lower value of MSE compared to the baselines across both the datasets. These results show the usefulness of \textit{MetaErr} for a regression application. Note that the graph of \textit{MetaErr} depicts an increasing trend with increasing DR, as we are plotting the error on the $y$-axis, and not the accuracy.

\subsubsection{Semantic Image Segmentation}
 
\hspace{.2in} \textbf{Datasets:} We used three datasets for the image segmentation experiment: $(i)$ \textbf{PASCAL-VOC} \cite{Pascal_dataset}; $(ii)$ \textbf{CityScapes} \cite{Cityscapes_dataset}; and $(iii)$ \textbf{Brain MRI} \cite{MRI_dataset}. 

\textbf{Comparison Baselines:} We used \textit{Random} and \textit{BALD} \cite{BALD_Paper} as the baselines in this setup. \textit{SR} was not used as it is challenging to come up with one posterior probability measure for a single image in this context. 

\textbf{Implementation Details:} \textbf{Base Model $\mathcal{F}$:} The U-Net architecture was used as the base model, as it is specifically designed for image segmentation. It incorporates multiple layers for up-sampling, down-sampling, and bottlenecks. Each of these layers is structured as a sequence of operations, commencing with two convolutional layers accompanied by \textit{BatchNorm} and \textit{ReLU} activation. These convolutional layers are defined with dimensions of $3 \times 3$, padding of $1$, and a stride of $1 \times 1$. The model was trained for $10$ epochs using the \textit{Adam} optimizer with a learning rate of $10^{-2}$. \textbf{Meta-Model $\mathcal{M}$:} We used the same meta model with the same training parameters as the Regression experiments. 

\begin{figure}[h]
	\centering
		\subfigure[PASCAL-VOC]{
          \label{fig_pascal}
          \includegraphics[trim = 1.3in 3.2in 1.7in 3.4in,clip,width=.31\textwidth]{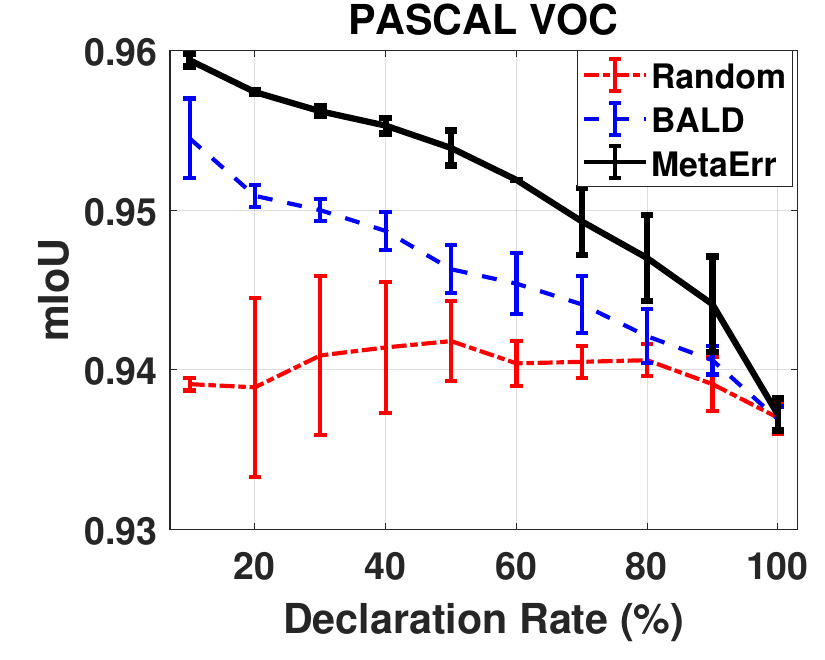}} 
     \hspace{.03in}    
     \subfigure[Cityscapes]{
          \label{fig_cityscapes}
          \includegraphics[trim = 1.3in 3.2in 1.7in 3.4in,clip,width=.31\textwidth]{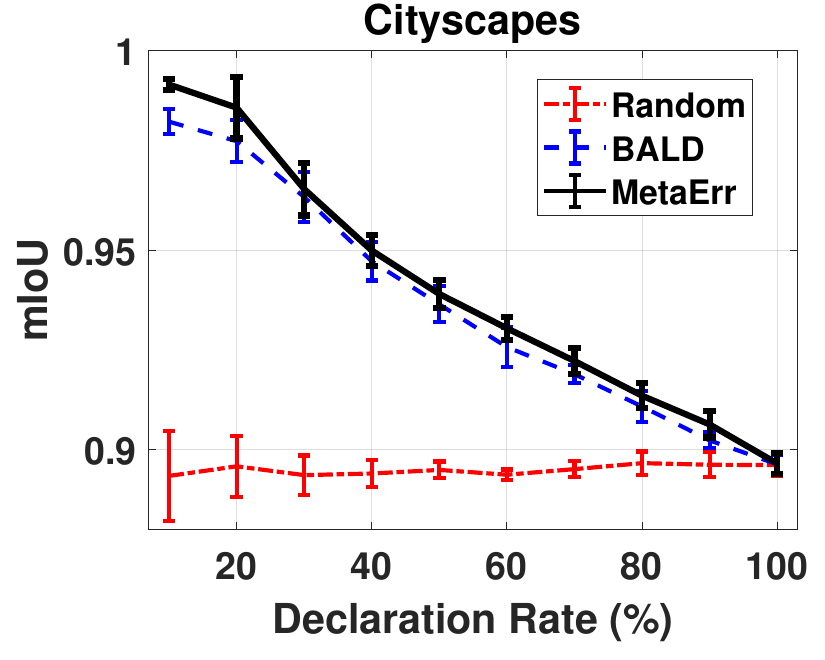}}  
     \hspace{.03in}    
     \subfigure[Brain MRI]{
          \label{fig_mri}
          \includegraphics[trim = 1.3in 3.2in 1.7in 3.4in,clip,width=.31\textwidth]{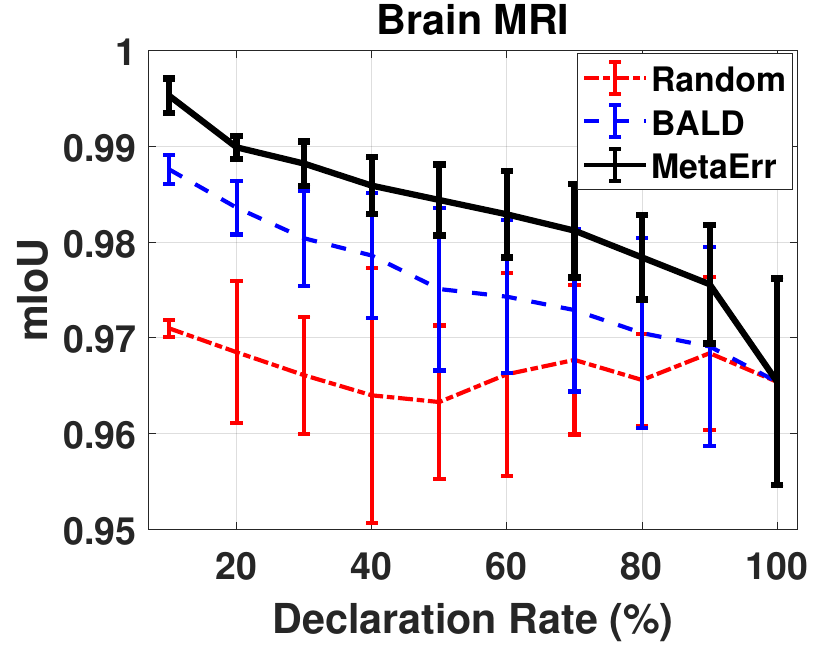}}       
		 		
	\caption{Performance of \textit{MetaErr} on semantic image sementation tasks. Best viewed in color.}
	\label{fig_seg_results}
\end{figure} 

\textbf{Results:} The results are depicted in Figure \ref{fig_seg_results}, where the $x$-axis denotes the DR and the $y$-axis denotes the mean Intersection-over-Union (mIoU). A similar pattern is evident in all the figures, with \textit{MetaErr} outperforming all the baselines consistently. At any given DR, it exhibits a higher mIoU than the baselines across all the three datasets. This shows that our framework can reliably identify the images which will be more accurately segmented by the base model. 

These results further corroborate the generalizability of \textit{MetaErr} across a variety of learning tasks.


\end{document}